\def\year{2022}\relax
\documentclass[letterpaper]{article} % DO NOT CHANGE THIS
\usepackage{aaai22}  % DO NOT CHANGE THIS
\usepackage{times}  % DO NOT CHANGE THIS
\usepackage{helvet}  % DO NOT CHANGE THIS
\usepackage{courier}  % DO NOT CHANGE THIS
\usepackage[hyphens]{url}  % DO NOT CHANGE THIS
\usepackage{graphicx} % DO NOT CHANGE THIS
\usepackage{natbib}  % DO NOT CHANGE THIS AND DO NOT ADD ANY OPTIONS TO IT
\usepackage{caption} % DO NOT CHANGE THIS AND DO NOT ADD ANY OPTIONS TO IT
\DeclareCaptionStyle{ruled}{labelfont=normalfont,labelsep=colon,strut=off} % DO NOT CHANGE THIS
\usepackage{algorithm}
\usepackage{algorithmic}

\usepackage{newfloat}
\usepackage{listings}
\usepackage{subfig}
\usepackage{multirow}
\usepackage[switch,mathlines]{lineno}
\usepackage{xcolor}
\usepackage{latexsym}
\usepackage{amsmath}
\usepackage{comment}
\usepackage{flexisym}

\floatstyle{ruled}
\newfloat{listing}{tb}{lst}{}
\floatname{listing}{Listing}
\title{Paradigm Shift in Language Modeling: Revisiting CNN for Modeling Sanskrit Originated Bengali and Hindi Language}

\author{
      Chowdhury Rafeed Rahman, \textsuperscript{\rm 1,2}
     MD. Hasibur Rahman, \textsuperscript{\rm 2}
     Mohammad Rafsan, \textsuperscript{\rm 2}
     Samiha Zakir, \textsuperscript{\rm 2}
     Mohammed Eunus Ali, \textsuperscript{\rm 1}
     Rafsanjani Muhammod \textsuperscript{\rm 2}
 }
 \affiliations{
 Bangladesh University of Engineering and Technology, \textsuperscript{\rm 1}
 United International University \textsuperscript{\rm 2}
 }

\begin{document}

\maketitle

\begin{abstract}
Though there has been a large body of recent works in language modeling (LM) for high resource languages such as English and Chinese, the area is still unexplored for low resource languages like Bengali and Hindi. We propose an end to end trainable memory efficient CNN architecture named \textit{CoCNN} to handle specific characteristics such as high inflection, morphological richness, flexible word order and phonetical spelling errors of Bengali and Hindi. In particular, we introduce two learnable convolutional sub-models at word and at sentence level that are end to end trainable. We show that state-of-the-art (SOTA) Transformer models including pretrained BERT do not necessarily yield the best performance for Bengali and Hindi. \textit{CoCNN} outperforms pretrained BERT with 16X less parameters, and it achieves much better performance than SOTA LSTM models on multiple real-world datasets. This is the first study on the effectiveness of different architectures drawn from three deep learning paradigms - Convolution, Recurrent, and Transformer neural nets for modeling two widely used languages, Bengali and Hindi.
\end{abstract}

% \noindent 
\section{Introduction} \label{intro}
\begin{figure*}[!htb]
    \begin{center}
    \includegraphics[width=1.2\columnwidth]{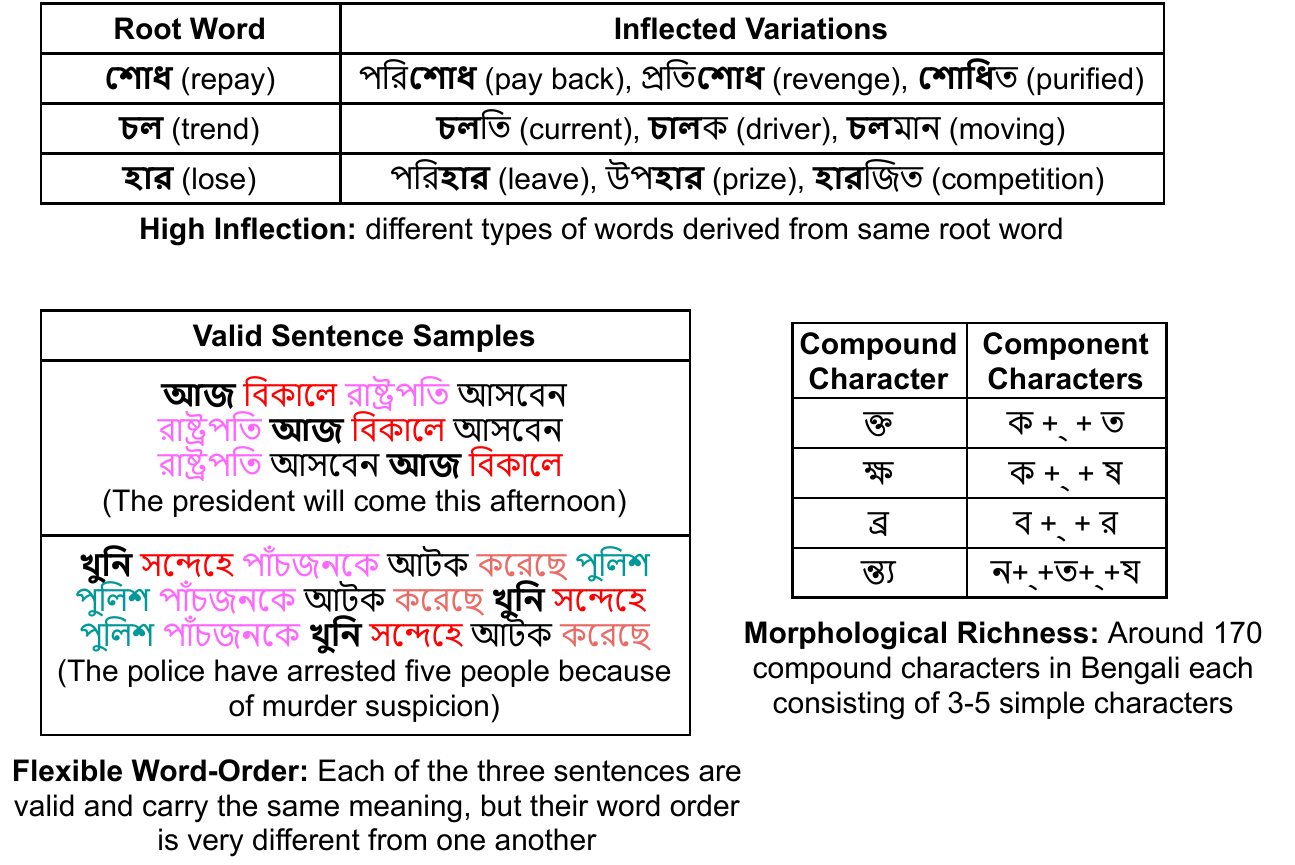}
    \end{center}
\caption{Bengali language unique characteristics}
\label{fig: Bengali}
\end{figure*}

Bengali and Hindi are the fourth and sixth most spoken language in the world, respectively. Both of these languages originated from Sanskrit \cite{intro6} and share some unique characteristics that include (i) high inflection, i.e., each root word may have many variations due to addition of different suffixes and prefixes, (ii) morphological richness, i.e., there are large number of compound letters, modified vowels and modified consonants, and (iii) flexible word-order, i.e., the importance of word order and their positions in a sentence are loosely bounded (Examples shown in Figure \ref{fig: Bengali}). Many other languages such as Nepali, Gujarati, Marathi, Kannada, Punjabi and Telugu also share these characteristics. Neural language models (LM) have shown great promise recently in solving several key NLP tasks such as word prediction and sentence completion in major languages such as English and Chinese \citep{LSTM1,LSTM_rel8,CNN1,intro7,intro8,intro9}. To the best of our knowledge, none of the existing study investigates the efficacy of recent LMs in the context of Bengali and Hindi. We conduct an in-depth analysis of major deep learning architectures for LM and propose an end to end trainable memory efficient CNN architecture to address the unique characteristics of Bengali and Hindi.

State-of-the-art (SOTA) techniques for LM can be categorized into three sub-domains of deep learning: (i) convolutional neural network (CNN) \citep{CNN1,CNN_rel3}  (ii) recurrent neural network \citep{LSTM2,LSTM3,LSTM4,LSTM5},  and (iii) Transformer attention network \citep{tr1,tr2,tr3,tr_rel4}. Long Short Term Memory (LSTM) based models, which are suitable for learning sequence and word order information, are not effective for modeling Bengali and Hindi due to their flexible word order characteristic. On the other hand, Transformers use dense layer based multi-head attention mechanism. They lack the ability to learn local patterns in sentence level, which in turn puts negative effect on modeling languages with loosely bound word order. Most importantly, neither LSTMs nor Transformers use any suitable measure to learn intra-word level local pattern necessary for modeling highly inflected and morphologically rich languages.   
%In contrast, contemporary CNN based architectures are good at local pattern learning. In fact, convolution operation is similar to attention mechanism as it puts more importance on certain patterns than others for providing us the correct output.

We observe that learning inter (flexible word order) and intra (high inflection and morphological richness) word local patterns is of paramount importance for Bengali and Hindi LM. To accommodate such characteristics, we design a novel CNN architecture, namely \textbf{Coordinated CNN (CoCNN)} that achieves SOTA performance with low training time. In particular, \textit{CoCNN} consists of two learnable convolutional sub-models: word level (\textit{Vocabulary Learner (VL)}) and sentence level (\textit{Terminal Coordinator (TC)}). \textit{VL} is designed for syllable pattern learning, whereas \textit{TC} serves the purpose of word coordination learning while maintaining positional independence, which suits the flexible word order of Bengali and Hindi. \textit{CoCNN} does not explicitly incorporate any self attention mechanism like Transformers; rather it relies on \textit{TC} for emphasizing on important word patterns. \textit{CoCNN} achieves significantly better performance than pretrained BERT for Bengali and Hindi LM with 16X less parameters. We further enhance \textit{CoCNN} by introducing skip connection and parallel convolution branches in \textit{VL} and \textit{TC}, respectively. This modified architecture (with negligible increase in parameter number) is named as \textit{CoCNN+}. We validate the effectiveness of \textit{CoCNN+} on a number of tasks that include next word prediction in erroneous setting, text classification, sentiment analysis and spell checking. \textit{CoCNN+} shows superior performance than contemporary LSTM based models and pretrained BERT.    

In summary, the contributions of this paper are as follows:
\begin{itemize}
    \item We propose an end to end trainable CNN architecture \textit{CoCNN} based on the coordination of two CNN sub-models. 
    \item We perform an in-depth analysis and comparison on different SOTA LMs in three paradigms: CNN, LSTM, and Transformer. With an extensive set of experiments, we show that CoCNN shows superior performance than LSTM and Transformer based models in spite of being memory efficient.
    \item We further show that simple modifications in \textit{CoCNN} can give us even more superior performance in terms of Bengali and Hindi language modeling and other downstream tasks like text classification and sentiment analysis.
    \item We show the potential of \textit{VL} sub-model of \textit{CoCNN+} as an effective spell checker for Bengali language.  
\end{itemize}

\section{Our Approach}

\begin{figure*}[!htb]
    \begin{center}
    \includegraphics[width=1.5\columnwidth]{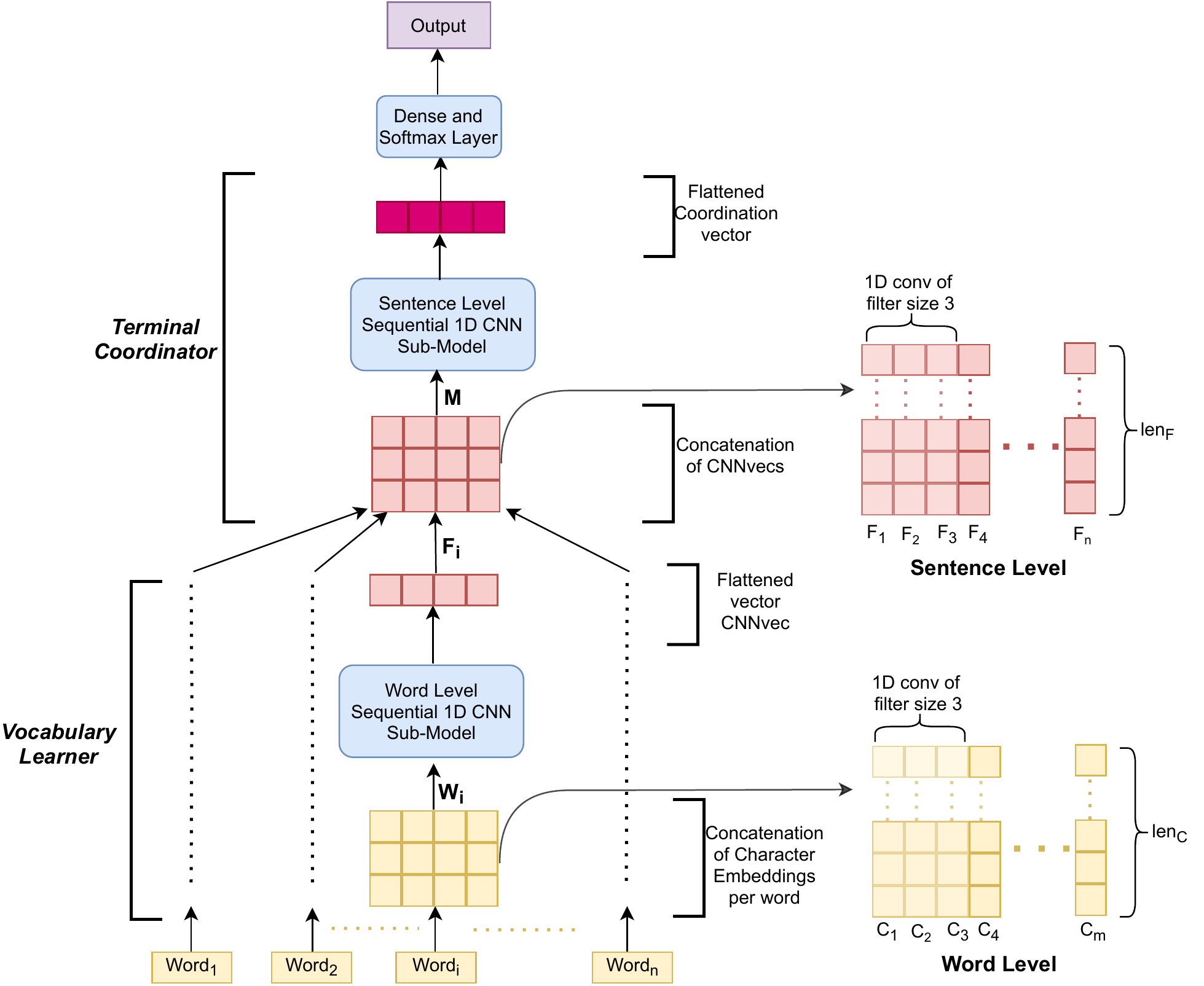}
    \end{center}
\caption{1D CNN based CoCNN architecture}
\label{fig: CNN}
\end{figure*}

Traditional CNN based approaches \citep{CNN1} represent the entire input sentence/ paragraph using a matrix of size $S_N \times S_V$, where $S_N$ and $S_V$ represent number of characters in the sentence/ paragraph and the character representation vector size, respectively. In such character based approach, the model does not have the ability to consider each word in the sentence as a separate entity. However, it is important to understand the contextual meaning of each word and to find out relationship among those words for sentence semantics understanding. Our proposed \textbf{Coordinated CNN (CoCNN)} is aimed to achieve this feat. Figure \ref{fig: CNN} illustrates \textit{CoCNN} that has two major components. \textit{Vocabulary Learner} component works at word level, while \textit{Terminal Coordinator} component works at sentence/ paragraph level. Both of these components are 1D CNN based sub-model at their core and are trained end-to-end.   

\subsection{\textit{Vocabulary Learner}} \label{vocab}
\textit{Vocabulary Learner (VL)} is used to transform each input word into a vector representation called \textit{CNNvec}. We represent each input word $Word_i$ by a matrix $W_i$. $W_i$ consists of $m$ vectors each of size $len_C$. These vectors $\vec{C_1}, \vec{C_2}, \ldots \vec{C_m}$ represent one hot vector of character $C_1, C_2, \ldots C_m$, respectively of $Word_i$.
Representation detail has been depicted in the bottom right corner of Figure \ref{fig: CNN}. 
Applying 1D convolution (\textit{conv}) layers on matrix $W_i$ helps in deriving key local patterns and sub-word information of $Word_i$. 1D \textit{conv} operation starting from vector $\vec{C_j}$ of $W_i$ using a \textit{conv} filter of size $K$ can be expressed as $1D\_Conv([\vec{C_j}, \vec{C_{j+1}}, \ldots \vec{C_{j+K-1}}], filter_K)$ function. The output of this function is a scalar value. While using a stride of $s$, the next \textit{conv} operation will start from vector $\vec{C_{j+s}}$ of $W_i$ and will provide us with another scalar value. Thus we get a vector as a result of applying one \textit{conv} filter on matrix $W_i$.
A \textit{conv} layer has multiple such \textit{conv} filters. After passing $W_i$ matrix through the first \textit{conv} layer we obtain feature matrix $W^1_i$. Passing $W^1_i$ through the second \textit{conv} layer provides us with feature matrix $W^2_i$. So, the $L^{th}$ \textit{conv} layer provides us with feature matrix $W^L_i$. \textit{VL} sub-model consists of such 1D \textit{conv} layers standing sequentially one after the other. \textit{Conv} layers near matrix $W_i$ are responsible for identifying key sub-word patterns of $Word_i$, while \textit{conv} layers further away focus on different combinations of these key sub-word patterns. Such word level local pattern recognition plays key role in identifying semantic meaning of a word irrespective of inflection or presence of spelling error. Each intermediate \textit{conv} layer output is batch normalized. The final \textit{conv} layer output matrix $W^L_i$ is flattened and formed into a vector $F_i$ of size $len_F$. $F_i$ is the \textit{CNNvec} representation of $Word_i$. We obtain \textit{CNNvec} representation from each of our input words in a similar fashion applying the same \textit{CNN} sub-model. 

\subsection{\textit{Terminal Coordinator}}
\textit{Terminal Coordinator (TC)} takes the \textit{CNNvecs} obtained from \textit{VL} as input and returns a single \textit{Coordination vector} as output which is used for final prediction. 
For $n$ words $Word_1, Word_2, \ldots Word_n$; we obtain $n$ such \textit{CNNvecs} $\vec{F_1}, \vec{F_2}, \ldots \vec{F_n}$, respectively. Each \textit{CNNvec} is of size $len_F$. Concatenating these \textit{CNNvecs} provide us with matrix $M$ (details shown in the middle right portion of Figure \ref{fig: CNN}).
Applying 1D \textit{conv} on matrix $M$ facilitates the derivation of key local patterns found in input sentence/ paragraph which is crucial for output prediction. 1D \textit{conv} operation starting from $\vec{F_i}$ using a \textit{conv} filter of size $K$ can be expressed as $1D\_Conv([\vec{F_i}, \vec{F_{i+1}}, \ldots \vec{F_{i+K-1}}], filter_K)$ function.  
The output of this function is a scalar value. A sequential 1D CNN sub-model with design similar to \textit{VL} (see Subsection \ref{vocab}) having different set of weights is employed on matrix $M$. \textit{Conv} layers near $M$ are responsible for identifying key word clusters, while \textit{conv} layers further away focus on different combinations of these key word clusters important for sentence or paragraph level local pattern recognition. The final output feature matrix obtained from the 1D CNN sub-model of \textit{TC} is flattened to obtain the \textit{Coordination vector}, a summary of important information obtained from the input word sequence in order to predict the correct output. 

\subsection{Attention to Patterns of Significance} \label{attention}

\begin{figure}[!htb]
    \begin{center}
    \includegraphics[width=0.8\columnwidth]{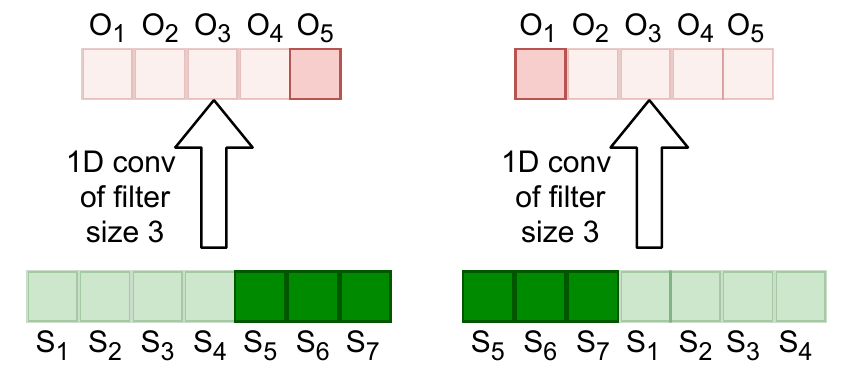}
    \end{center}
\caption{1D convolution as a form of attention}
\label{fig: attention}
\end{figure}

There is no explicit attention mechanism in \textit{CoCNN} unlike self attention based Transformers \citep{tr2,tr3} or weighted attention based LSTMs \citep{attn1,attn2}. 
Attention mechanism is important for obtaining the importance of each input word in terms of output prediction. Figure \ref{fig: attention} demonstrates an over simplified notion of how 1D \textit{conv} implicitly imposes attention on a sequence containing 7 entities $S_1, S_2, \ldots S_7$. After employing a 1D \textit{conv} filter of size 3 on this sequence representation, we obtain a vector containing values $O_1, O_2, \ldots O_5$, where we assume $O_5$ and $O_1$ to be the maximum for the left and the right figure, respectively. $O_5$ (left figure) is obtained by convolving input entity $S_5$, $S_6$ and  $S_7$ situated at the end of input sequence. We can say that these three input entities have been paid more attention to by our \textit{conv} filter than the other 4 entities. In the right figure, $S_5$, $S_6$ and  $S_7$ are situated at the beginning of the input sequence. Our \textit{conv} filter gives similar importance to this pattern irrespective of its position change. Such positional independence of important patterns helps in Bengali and Hindi LM where input words are loosely bound and words themselves are highly inflected. In CoCNN, such attention is imposed on characters of each word during \textit{CNNvec} representation generation of that word using \textit{VL}, while similar type of attention is imposed on words of our input sentence/ paragraph while obtaining \textit{Coordination vector} using  \textit{TC}.  

\begin{comment}
\subsection{Positive Effect of Batch Normalization} \label{batch}
We implement batch normalization on all \textit{conv} layer outputs of both \textit{VL} and \textit{TC} of proposed \textit{CoCNN} architecture while performing mini batch training. Let us take a single \textit{conv} output for example. 
Suppose, after passing sample no. $i$ of a mini batch, we receive $out^\textprime_i$ as the preliminary version of this output value. We modify $out^\textprime_i$ to get the final version $out_i$ as follows:
\begin{center}
$y_{i} = \frac{out^\textprime_i - mean_B}{\sqrt{var_B + \epsilon}}$
\end{center}
\begin{center}
$out_{i} = w \times y_{i} + b$
\end{center}

Here, $var_B$ and $mean_B$ denote variance and mean of this output value for the current batch of samples. $w$ and $b$ are learnable parameters kept for reducing internal covariate shift in terms of this output value over all batches. Since we have a wide variety of words and various types of semantically meaningful sentences at our disposal as training samples, batch normalized output values provides us with three types of benefits over raw output values - (1) faster training, (2) less bias towards initial weights and (3) automatic regularization effect \citep{batch1,batch2}. 
\end{comment}

\subsection{Extending CoCNN}
\begin{figure}[!htb]
    \begin{center}
    \includegraphics[width=0.75\columnwidth]{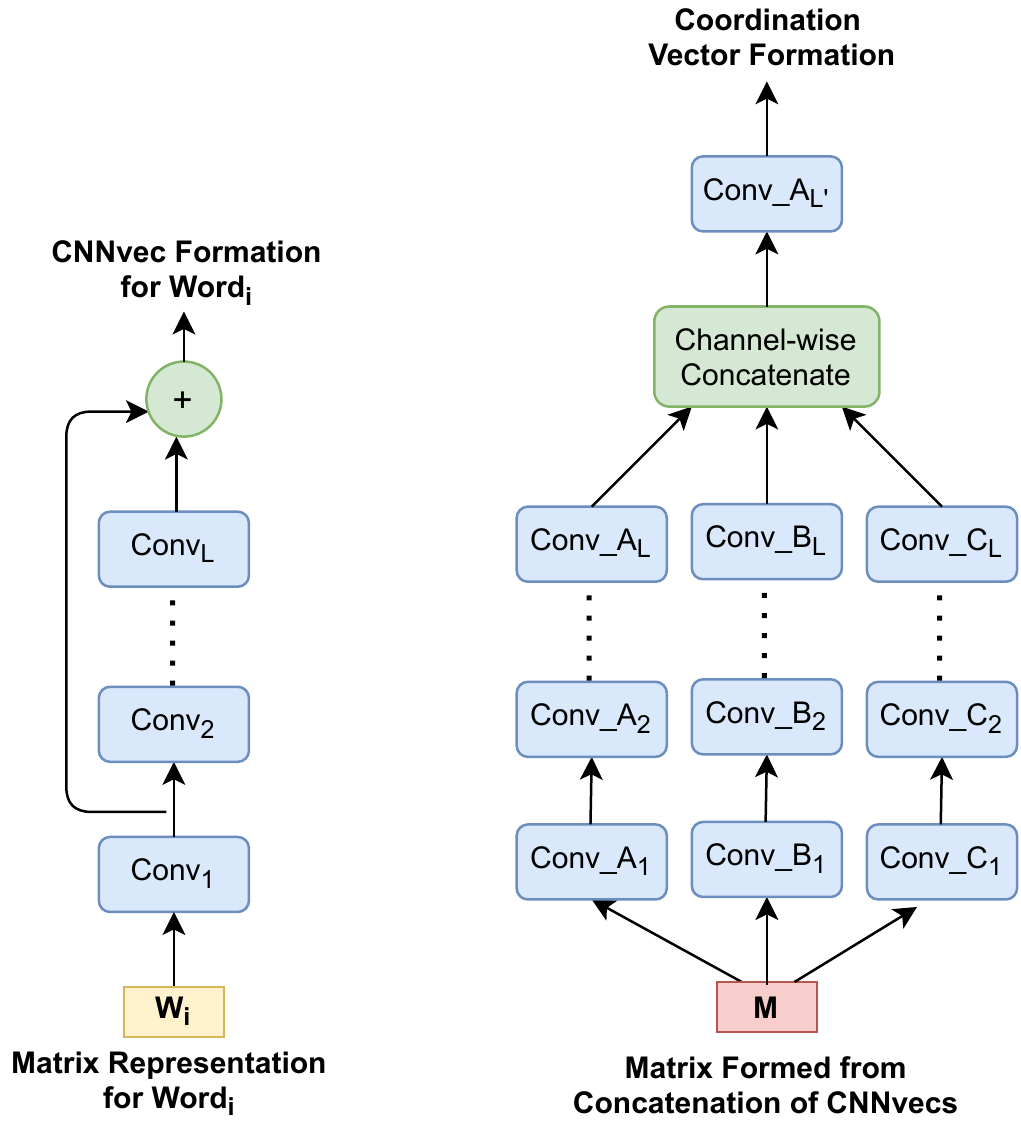}
    \end{center}
\caption{\textit{CoCNN+} architecture with its modified \textit{VL} (left) and \textit{TC} (right). $Conv_L$ means $L^{th}$ \textit{conv} layer, whereas \textit{Conv\_A} means a \textit{conv} layer with filter size A.}
\label{fig: CoCNN+}
\end{figure}

We perform two simple modifications in \textit{CoCNN} to form \textit{CoCNN+} architecture with minimal increase in parameter number (see Figure \ref{fig: CoCNN+}). \\
\textbf{First}, we modify the CNN sub-model of \textit{VL}. We add the output feature matrix of the first \textit{conv} layer $Conv_1$ with the output feature matrix of the last \textit{conv} layer $Conv_L$. We pass the resultant feature matrix on to subsequent layers (same as \textit{CoCNN}) for \textit{CNNvec} formation of $Word_i$. Such modification helps in two cases - (i) it eliminates the gradient vanishing problem of the first \textit{conv} layer of \textit{VL} and (ii) it gives \textit{CNNvec} access to both low level and high level features of the corresponding input word.\\
\textbf{Second}, we modify the CNN sub-model of \textit{TC} by passing matrix $M$ simultaneously to three 1D CNN branches. The \textit{conv} filter sizes of the left, middle and right branches are $A$, $B$ and $C$, respectively; where, $A<B$ and $B<C$. The outputs from the three branches are concatenated channel-wise and are then passed on to the final \textit{conv} layer having filter size $A$. The output feature matrix is passed on to subsequent layers (same as \textit{CoCNN}) for \textit{Coordination vector} formation. Multiple \textit{conv} branches with different filter sizes help in learning both short and long range local patterns, especially when the input sentence or document is long.       

\section{Experimental Setup}
\subsection{Dataset Specifications} \label{Data}
Bengali dataset consists of articles from online public news portals such as Prothom-Alo \citep{data1}, BDNews24 \citep{data2} and Nayadiganta \citep{data3}. The articles encompass domains such as politics, entertainment, lifestyle, sports, technology and literature. The Hindi dataset consists of Hindinews \citep{data6}, Livehindustan \citep{data4} and Patrika \citep{data5} newspaper articles available open source in Kaggle encompassing similar domains. Nayadiganta (Bengali) and Patrika (Hindi) datasets have been used only as independent test sets. Detailed statistics of the datasets are provided in Table \ref{Table:Ds}. Top words have been selected such that they cover at least 90\% of the dataset. For each Bengali dataset, we have created a new version of the dataset by incorporating spelling errors using a probabilistic error generation algorithm \citep{intro2}, which enables us to test the effectiveness of LMs for erroneous datasets.
%We us as e a probability based algorithm for realistic spelling error generation in our Bengali corpus \citep{intro2} as Bengali writing using English Qwerty keyboard is error prone in practice.     
\begin{table*}[!htb]
\centering
\begin{tabular}{|c|c|c|c|c|c|}
\hline
\textbf{Datasets} & \textbf{\begin{tabular}[c]{@{}c@{}}Unique\\ Word No.\end{tabular}} & \textbf{\begin{tabular}[c]{@{}c@{}}Unique\\ Character No.\end{tabular}} & \textbf{\begin{tabular}[c]{@{}c@{}}Top\\ Word No.\end{tabular}} & \textbf{\begin{tabular}[c]{@{}c@{}}Training\\ Sample No.\end{tabular}} & \textbf{\begin{tabular}[c]{@{}c@{}}Validation\\ Sample No.\end{tabular}} \\ \hline
Prothom-Alo       & 260 K                                                              & 75                                                                      & 13 K                                                            & 5.9 M                                                                  & 740 K                                                                    \\ \hline
BDNews24          & 170 K                                                              & 72                                                                      & 14 K                                                            & 2.9 M                                                                  & 330 K                                                                    \\ \hline
Nayadiganta       & 44 K                                                               & 73                                                                      & \_                                                              & \_                                                                     & 280 K                                                                    \\ \hline
Hindinews         & 37 K                                                               & 74                                                                      & 5.5 K                                                           & 87 K                                                                   & 10 K                                                                     \\ \hline
Livehindustan     & 60 K                                                               & 73                                                                      & 4.5 K                                                           & 210 K                                                                  & 20 K                                                                     \\ \hline
Patrika           & 28 K                                                               & 73                                                                      & \_                                                              & \_                                                                     & 307 K                                                                    \\ \hline
\end{tabular}
\caption{Dataset details (K and M denote $10^3$ and $10^6$ multiplier, respectively)}
\label{Table:Ds}
\end{table*}

\subsection{Performance Metric}
We use perplexity (PPL) to assess the performance of the models for next word prediction task. Suppose, we have sample inputs $I_1, I_2, \dots, I_n$ and our model provides probability values $P_{1}, P_{2}, \ldots, P_{n}$, respectively for their ground truth output tokens. Then the PPL score of our model for these samples can be computed as:
\begin{center}
    $PPL = \exp(-\frac{1}{n}\sum_{i=1}^{n} \ln(P_{i}))$
\end{center}
PPL as a metric emphasizes on a model's ability to understand a language instead of emphasizing on predicting the ground truth next word as output.  
For text classification and sentiment analysis, we use \textit{accuracy} and \textit{F1 score} as our performance metric.

\subsection{Model Optimization}
For model optimization, we use SGD optimizer with a learning rate of 0.001 while constraining the norm of the gradients to below 5 for exploding gradient problem elimination. We use Categorical Cross-Entropy loss for model weight update and dropout \citep{exp1} with probability 0.3 between the dense layers for regularization. We use Relu (Rectified Linear Unit) as hidden layer activation function. We use a batch size of 64. As we apply batch normalization on CNN intermediate outputs, we do not use any other regularization effect such as dropout on these layers \citep{batch3}. 

We use Anaconda 3 with Python 3.8 version and Tensorflow 2.6.0 framework for our implementation.   
We use two GPU servers for training our models: (i) 12 GB Nvidia Titan Xp GPU, Intel(R) Core(TM) i7-7700 CPU (3.60GHz) processor model (ii) 32 GB RAM with 8 cores 24 GB Nvidia Tesla K80 GPU, Intel(R) Xeon(R) CPU (2.30GHz) processor model

\subsection{CoCNN Hyperparameters}
Our proposed \textit{CoCNN} architecture has two main components. We specify the details of each of them in this subsection.

\subsubsection{\textit{Vocabulary Learner} Details}
\textit{Vocabulary Learner} sub-model consists of a character level embedding layer producing a 40 size vector from each character, then four consecutive layers each consisting of 1D convolution (batch normalization and Relu activation in between each pair of convolution layers) and finally, a 1D global maxpooling in order to obtain \textit{CNNvec} representation from each input word. The four 1D convolution layers consist of $(32, 2), (64, 3), (64, 3), (128, 4)$ convolution, respectively. Here the first and second element of each tuple denote number of convolution filters and kernel size, respectively. As we can see, the filter size and number of filters of the convolution layers are monotonically increasing as architecture depth increases. It is because deep convolution layers need to learn the combination of various low level features which is a more difficult task compared to the task of shallow layers that include extraction of low level features.

\subsubsection{\textit{Terminal Coordinator} Details}
The \textit{Terminal Coordinator} sub-model used in \textit{CoCNN} architecture uses six convolution layers which consist of $(32, 2), (64, 3), (64, 3), (96, 3), (128, 4), (196, 4)$ convolution. Its design is similar to that of \textit{Vocabulary Learner} sub-model. The final output feature matrix obtained from this CNN sub-model is flattened to get the \textit{Coordination vector}. After passing this vector through a couple of dense layers, we use \textit{Softmax} activation function at the final output layer to get the predicted output.

\subsection{CoCNN+ Hyperparameters}
The CNN sub-model of \textit{Vocabulary Learner} in \textit{CoCNN+} is the same as \textit{CoCNN} except for one aspect (see Figure \ref{fig: CoCNN+}) - we change the first convolution layer to have 128 filters of size 2 instead of 32 filters. This is done to respect the matrix dimensionality during skip connection based addition. 

Instead of providing a sequential 1D CNN sub-model in \textit{Terminal Coordinator}, we provide three parallel branches each consisting of four convolution layers (see Figure \ref{fig: CoCNN+}) where the filter numbers are 32, 64, 96 and 128. The filter size of the leftmost, middle and the rightmost branch are 3, 5 and 7, respectively. All convolution operations are dimension preserving through the use of padding. The feature matrices of all three of these branches are concatenated channel-wise and finally, this concatenated matrix is passed on to a final convolution layer with 196 filters of size 3. 

\subsection{Hardware Specifications}
We use Python 3.8 and Tensorflow 2.6.0 package for our implementation.   
We use three GPU servers for training our models. Their specifications are as follows:
\begin{itemize}
    \item 12 GB Nvidia Titan Xp GPU, Intel(R) Core(TM) i7-7700 CPU (3.60GHz) processor model, 32 GB RAM with 8 cores
    \item 24 GB Nvidia Tesla K80 GPU, Intel(R) Xeon(R) CPU (2.30GHz) processor model, 12 GB RAM with 2 cores
    \item 8 GB Nvidia RTX 2070 GPU, Intel(R) Core(TM) CPU (2.20GHz) processor model, 32 GB RAM with 7 cores
\end{itemize}

\section{Results and Discussion}\label{Results}

\begin{figure*}[!htb]
 \centering
    \subfloat[CNN paradigm\label{fig: CNN_comp}]
    {\includegraphics[width=2.25in]{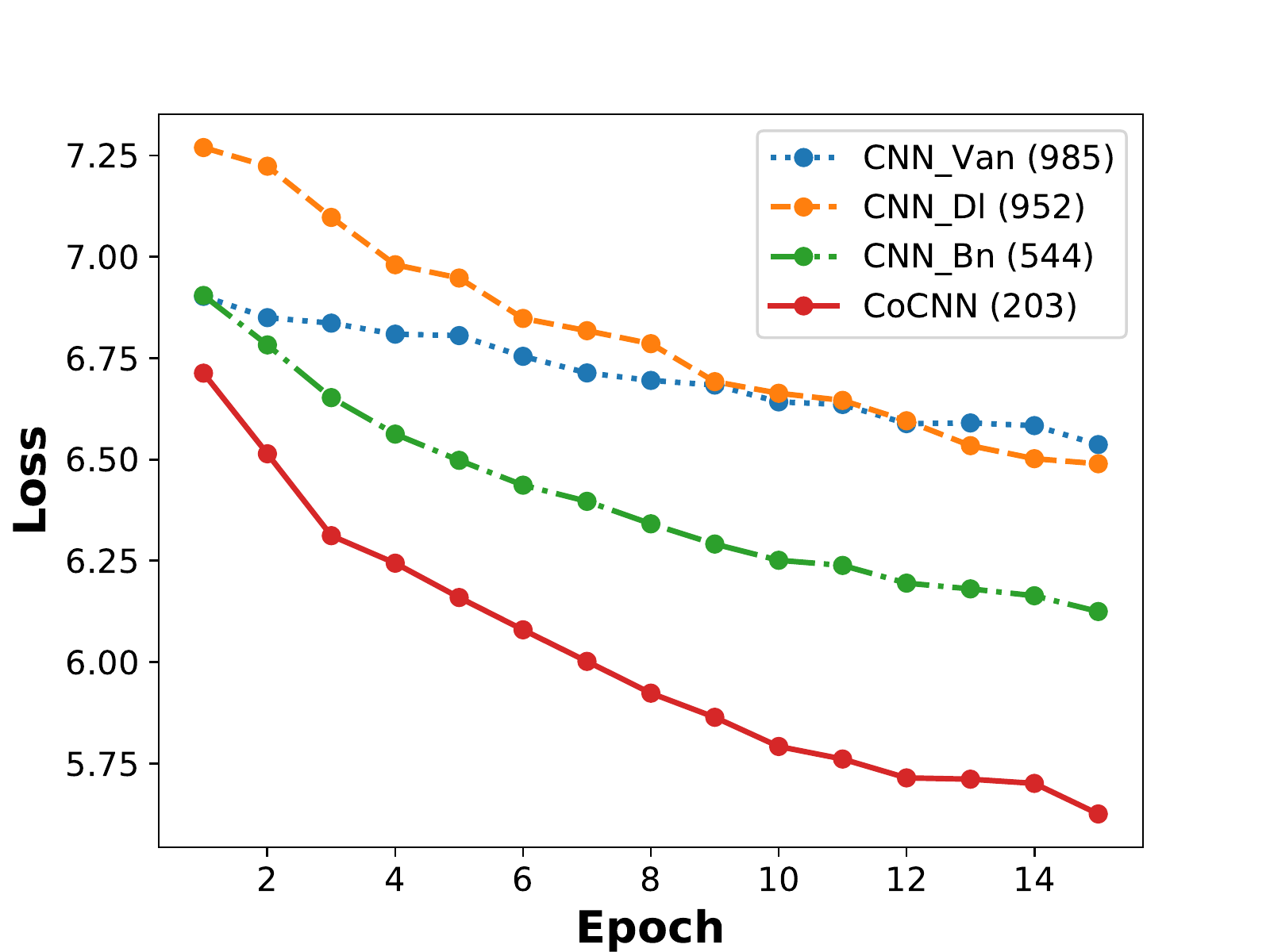}}
     \subfloat[LSTM paradigm\label{fig: LSTM_comp}]
     {\includegraphics[width=2.25in]{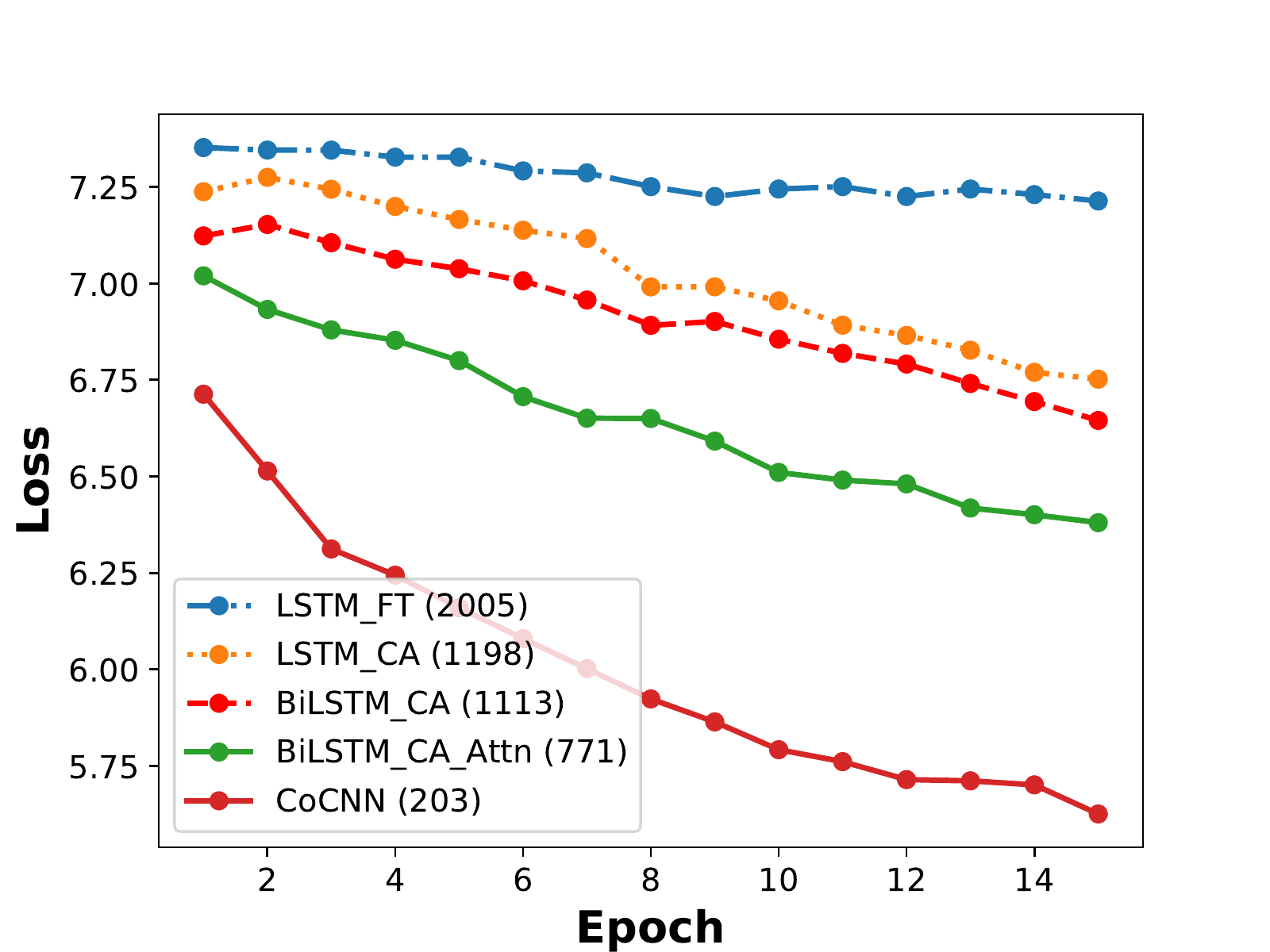}} 
     \subfloat[Transformer paradigm\label{fig: Trans_comp}]
     {\includegraphics[width=2.25in]{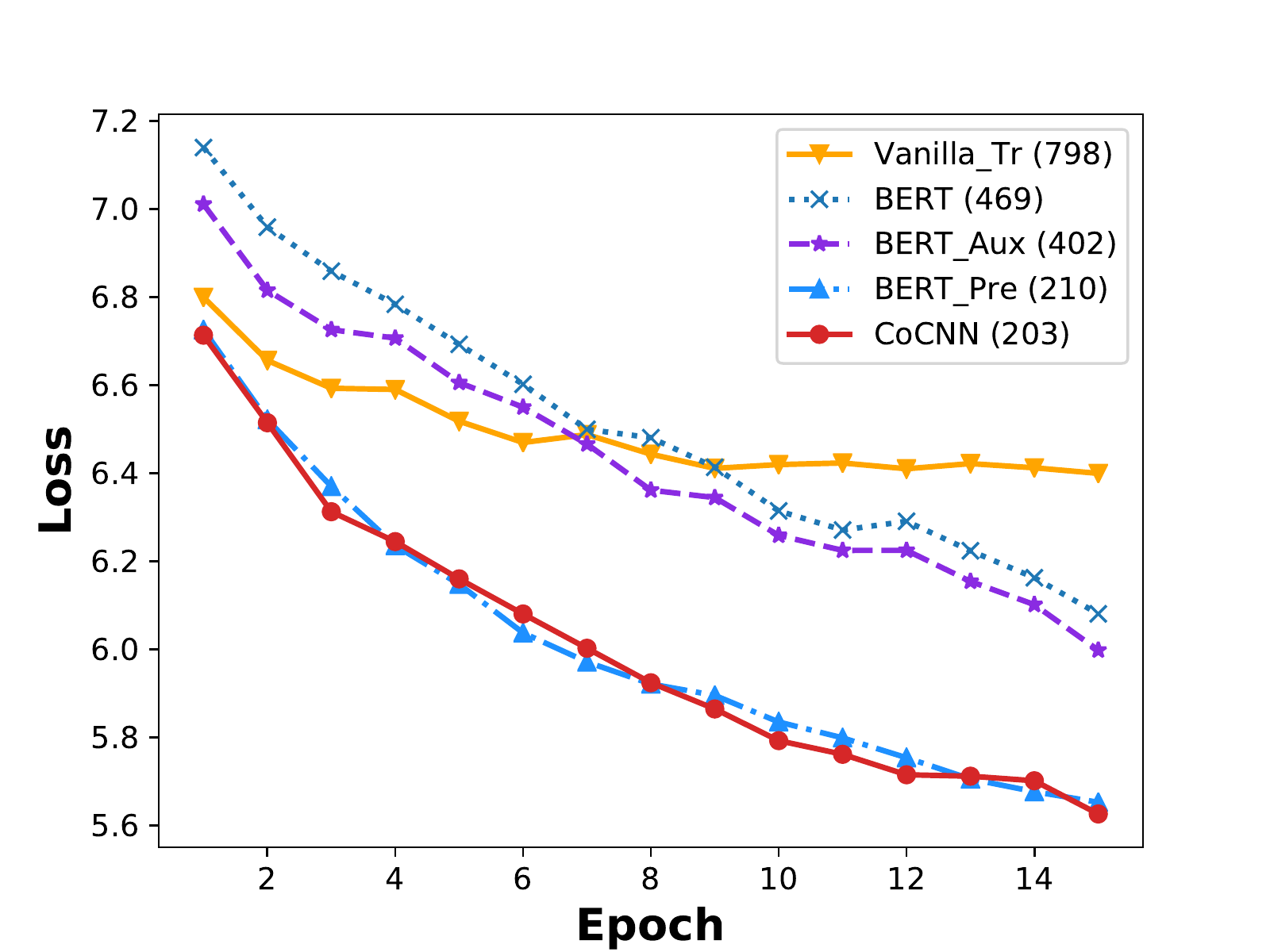}} 
\caption{Comparing \textit{CoCNN} with SOTA architectures from CNN, LSTM and Transformer paradigm on Prothom-Alo validation set. The score shown beside each model name denotes that model's PPL score on Prothom-Alo validation set after 15 epochs of training. Note that this dataset contains synthetically generated spelling errors.}
\label{fig: paradigms}
\end{figure*}

\subsection{Comparing \textit{CoCNN} with Other CNNs}
We compare \textit{CoCNN} with three other CNN-based baselines (see Figure \ref{fig: CNN_comp}). \textit{CNN\_Van} is a simple sequential 1D CNN model of moderate depth \citep{CNN1}. It considers the full input sentence/ paragraph as a matrix. The matrix consists of character representation vectors. \textit{CNN\_Dl} uses dilated \textit{conv} in its CNN layers which allows the model to have a larger field of view \citep{CNN2}. Such change in \textit{conv} strategy shows slight performance improvement. \textit{CNN\_Bn} has the same setting as of \textit{CNN\_Van}, but uses batch normalization on intermediate \textit{conv} layer outputs. Such measure shows significant performance improvement in terms of loss and PPL score. Proposed \textit{CoCNN} surpasses the performance of \textit{CNN\_Bn} by a wide margin. We believe that the ability of \textit{CoCNN} to consider each word of a sentence as a separate meaningful entity is the reason behind this drastic improvement.

\subsection{Comparing \textit{CoCNN} with SOTA LSTMs}
We compare \textit{CoCNN} with four LSTM-based models (see Figure \ref{fig: LSTM_comp}). Two LSTM layers are stacked on top of each other in all four of these models. We do not compare with LSTM models that use \textit{Word2vec} \cite{res1} representation as this representation requires fixed size vocabulary. In spelling error prone setting, vocabulary size is theoretically infinite. We start with \textit{LSTM\_FT}, an architecture using sub-word based \textit{FastText}  representation \citep{LSTM1,LSTM2}. Character aware learnable layers per LSTM time stamp form the new generation of SOTA LSTMs \citep{LSTM3,LSTM4,LSTM5,LSTM6}. \textit{LSTM\_CA} acts as their representative by introducing variable size parallel \textit{conv} filter output concatenation as word representation. The improvement over \textit{LSTM\_FT} in terms of PPL score is almost double. Instead of unidirectional many to one LSTM, we introduce bidirectional LSTM in \textit{LSTM\_CA} to form \textit{BiLSTM\_CA} which shows slight performance improvement. We introduce Bahdanu attention \citep{attn1} on \textit{BiLSTM\_CA} to form \textit{BiLSTM\_CA\_Attn} architecture. Such measure shows further performance boost. \textit{CoCNN} shows almost four times improvement in PPL score compared to \textit{BiLSTM\_CA\_Attn}. If we compare Figure \ref{fig: LSTM_comp} and \ref{fig: CNN_comp}, we can see that CNNs perform relatively better than LSTMs in general for Bengali LM.  LSTMs have a tendency of learning sequence order information which imposes positional dependency. Such characteristic is unsuitable for Bengali and Hindi with flexible word order.  

\subsection{Comparing \textit{CoCNN} with SOTA Transformers}
We compare \textit{CoCNN} with four Transformer-based models (see Figure \ref{fig: Trans_comp}). We use popular \textit{FastText} word representation with all compared transformers. Our comparison starts with \textit{Vanilla\_Tr}, a single Transformer encoder (similar to the Transformer designed by \citet{tr2}). In \textit{BERT}, we stack 12 transformers on top of each other where each Transformer encoder has more parameters than  
the Transformer of \textit{Vanilla\_Tr} \citep{BERT,tr3}. \textit{BERT} with its large depth and enhanced encoders almost double the performance shown by \textit{Vanilla\_Tr}. We do not pretrain this \textit{BERT} architecture. We follow the Transformer architecture designed by \citet{tr1} and introduce auxiliary loss after the Transformer encoders situated near the bottom of the Transformer stack of \textit{BERT} to form \textit{BERT\_Aux}. Introduction of such auxiliary losses show moderate improvement of performance. \textit{BERT\_Pre} is the pretrained version of \textit{BERT}. We follow the word masking based pretraining scheme of \citet{RoBERT}.  The Bengali pretraining corpus consists of Prothom Alo \citep{data1} news articles dated from 2014-2017 and BDNews24 \citep{data2} news articles dated from 2015-2017. The performance of \textit{BERT} jumps up more than double when such pretraining is applied. \textit{CoCNN} without utilizing any pretraining achieves marginally better performance than \textit{BERT\_Pre}. Unlike Transformer encoders, \textit{conv} imposes attention with a view to extracting important patterns from the input to provide the correct output (see Subsection \ref{attention}). Furthermore, \textit{VL} of \textit{CoCNN} is suitable for deriving semantic meaning of each input word in highly inflected and error prone settings. 

\subsection{Comparing \textit{BERT\_Pre}, \textit{CoCNN} and \textit{CoCNN+}}
\begin{figure}[!htb]
 \centering
     \subfloat[Plot on Bengali dataset\label{fig: prothomAlo_ds}]{\includegraphics[width=2.1in]{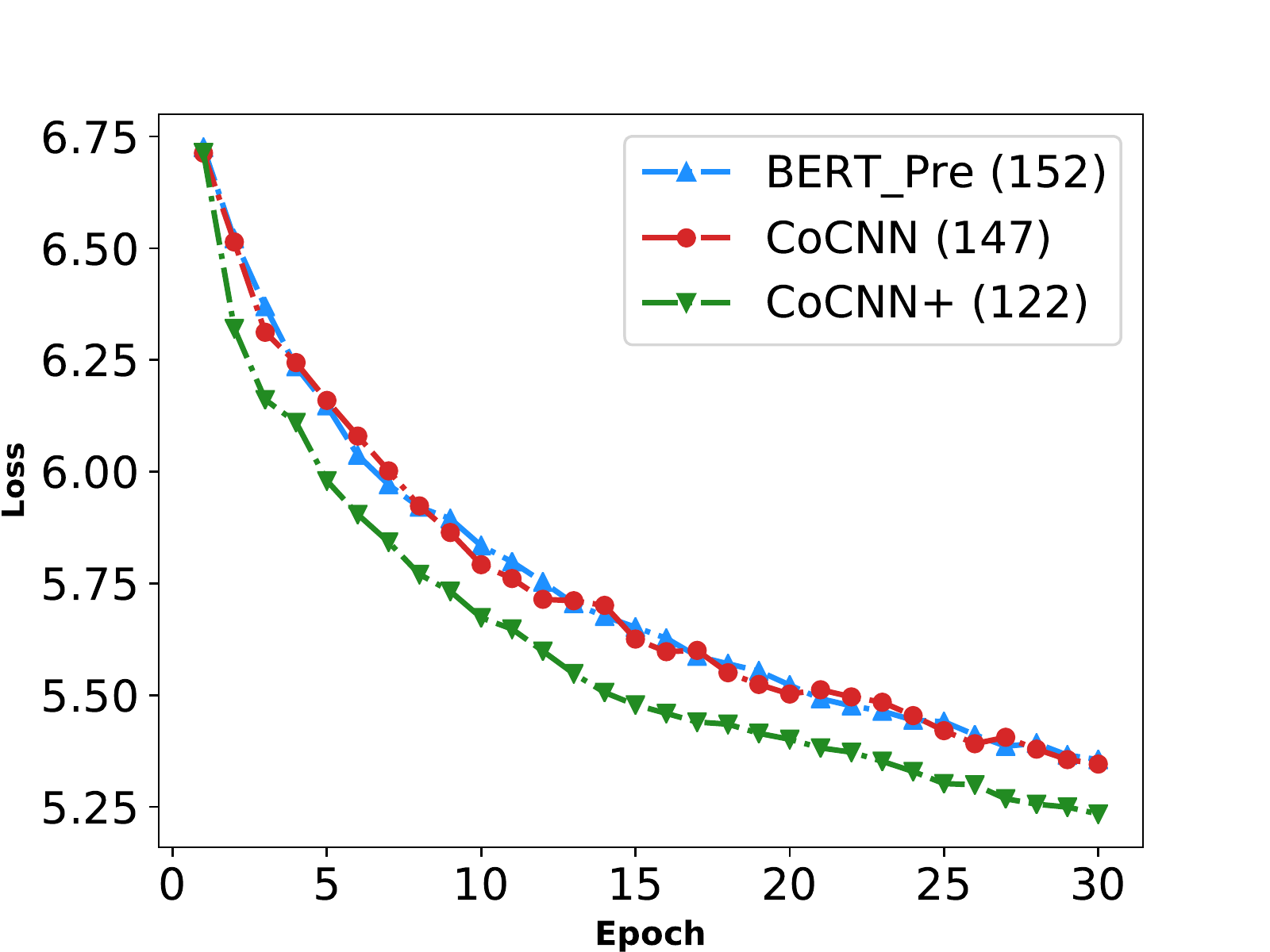}}
     
     \subfloat[Plot on Hindi dataset\label{fig: Hindi_comp}]{\includegraphics[width=2.1in]{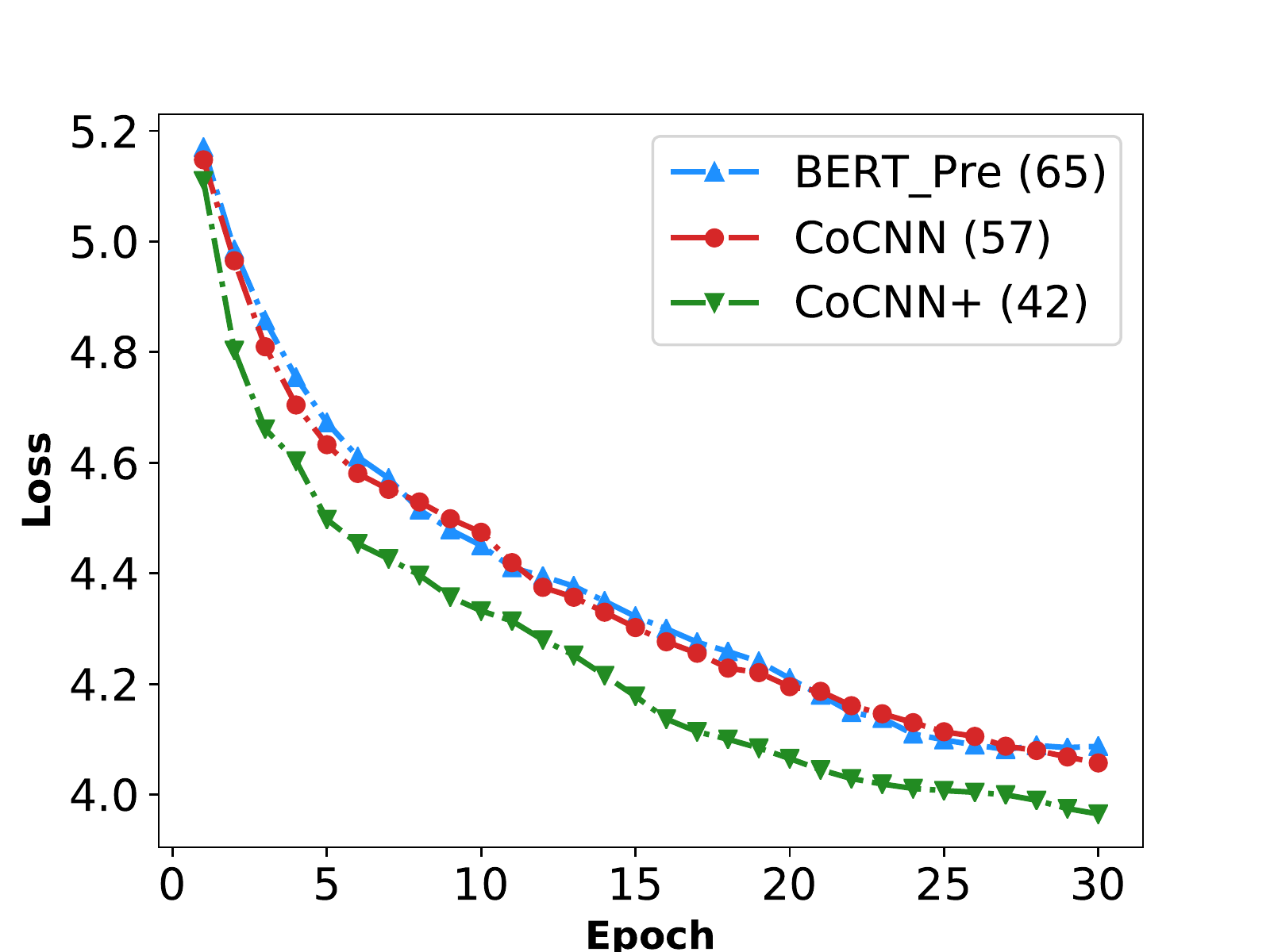}} 
\caption{Comparing \textit{BERT\_Pre}, \textit{CoCNN} and \textit{CoCNN+} on Bengali (Prothom-Alo) and Hindi (Hindinews and Livehindustan merged) validation set. The score shown beside each model name denotes that model's PPL score after 30 epochs of training on corresponding training set.}
\label{fig: best}
\end{figure}

\begin{table}[!htb]
\centering
\begin{tabular}{|c|c|c|c|c|}
\hline
\textbf{Datasets}                                                                    & \textbf{Error?}     & \textbf{\begin{tabular}[c]{@{}c@{}}BERT\_\\ Pre\end{tabular}} & \textbf{\begin{tabular}[c]{@{}c@{}}Co-\\ CNN\end{tabular}} & \textbf{\begin{tabular}[c]{@{}c@{}}Co-\\ CNN+\end{tabular}} \\ \hline
\multirow{2}{*}{Prothom-Alo}                                                         & Yes                 & 152                                                           & 147                                                        & 122                                                         \\ \cline{2-5} 
                                                                                     & No                  & 117                                                           & 114                                                        & 99                                                          \\ \hline
\multirow{2}{*}{BDNews24}                                                            & Yes                 & 201                                                           & 193                                                        & 170                                                         \\ \cline{2-5} 
                                                                                     & No                  & 147                                                           & 141                                                        & 123                                                         \\ \hline
\multirow{2}{*}{\begin{tabular}[c]{@{}c@{}}Hindinews +\\ Livehindustan\end{tabular}} & \multirow{2}{*}{No} & \multirow{2}{*}{65}                                           & \multirow{2}{*}{57}                                        & \multirow{2}{*}{42}                                         \\
                                                                                     &                     &                                                               &                                                            &                                                             \\ \hline
\multirow{2}{*}{Nayadiganta}                                                         & Yes                 & 169                                                           & 162                                                        & 143                                                         \\ \cline{2-5} 
                                                                                     & No                  & 136                                                           & 133                                                        & 118                                                         \\ \hline
Patrika                                                                              & No                  & 67                                                            & 57                                                         & 44                                                          \\ \hline
\end{tabular}
\caption{Comparing (PPL) \textit{BERT\_Pre}, \textit{CoCNN} and \textit{CoCNN+}}
\label{tab: ppl_comp}
\end{table}

\textit{BERT\_Pre} is the only model showing performance close to \textit{CoCNN} in terms of validation loss and PPL score (see Figure \ref{fig: paradigms}). We compare these two models with \textit{CoCNN+}. we train the models for 30 epochs on several Bengali and Hindi datasets and obtain their PPL scores on corresponding validation sets (training and validation set were split at 80\%-20\% ratio). Bengali datasets include Prothom-Alo, BDNews24; while Hindi dataset includes Hindinews, Livehindustan. We use Nayadiganta and Patrika dataset for Bengali and Hindi independent test set, respectively. The Hindi pretraining corpus consists of Hindi Oscar Corpus \citep{Oscar}, preprocessed Wikipedia articles \citep{Wiki}, HindiEnCorp05 dataset \citep{encorp} and WMT Hindi News Crawl data \citep{WMT}. From the graphs of Figure \ref{fig: best} and PPL score comparison Table \ref{tab: ppl_comp}, it is evident that \textit{CoCNN} marginally outperforms its nemesis \textit{BERT\_Pre} in all cases, while \textit{CoCNN+} outperforms both \textit{CoCNN} and \textit{BERT\_Pre} by a significant margin. There are 8 sets of PPL scores in Table \ref{tab: ppl_comp} for the three models on eight different dataset settings. We use these scores to perform one tailed paired t-test in order to determine whether the reduction of PPL score seen in \textit{CoCNN} compared to \textit{BERT\_Pre} is statistically significant when P-value threshold is set to 0.05. The statistical test shows that the improvement is significant. Similarly, \textit{CoCNN+} significantly outperforms \textit{CoCNN} in terms of achieved PPL scores. Number of parameters of \textit{BERT\_Pre}, \textit{CoCNN} and \textit{CoCNN+} are 74 M, 4.5 M and 4.8 M, respectively. Though the parameter number of \textit{CoCNN+} and \textit{CoCNN} is close, \textit{CoCNN+} has 15X fewer parameters than \textit{BERT\_Pre}.

\begin{comment}
\begin{table}[!htb]
\begin{tabular}{|c|c|c|}
\hline
\textbf{Model} & \textbf{Parameter} & \textbf{\begin{tabular}[c]{@{}c@{}}Training time\\ / epoch\end{tabular}} \\ \hline
BERT\_Pre  & 74 M               & 402 min                                                                  \\ \hline
CoCNN          & 4.5 M              & 40 min                                                                   \\ \hline
\end{tabular}
\caption{Comparing \textit{BERT\_Pre} and \textit{CoCNN} in terms of efficiency}
\label{tab: efficient}
\end{table}
\end{comment}

\subsection{Comparison in Downstream Tasks}
We first show the performance comparison of \textit{CoCNN+} with \textit{BERT\_Pre} in three downstream tasks. Then we present the performance of one of our key components \textit{VL} for spell checking task.
\begin{table}[!htb]
\centering
\begin{tabular}{|c|c|c|c|}
\hline
\textbf{Dataset}                                                                      & \textbf{Metric} & \textbf{BERT\_Pre} & \textbf{CoCNN+} \\ \hline
\multirow{2}{*}{\textbf{\begin{tabular}[c]{@{}c@{}}Question\\ Classify\end{tabular}}} & Acc             & 91.3\%                 & 93.7\%          \\ \cline{2-4} 
                                                                                      & F1              & 0.905                  & 0.926           \\ \hline
\multirow{2}{*}{\textbf{\begin{tabular}[c]{@{}c@{}}Product\\ Review\end{tabular}}}    & Acc             & 84.4\%                 & 86.2\%          \\ \cline{2-4} 
                                                                                      & F1              & 0.841                  & 0.86            \\ \hline
\multirow{2}{*}{\textbf{\begin{tabular}[c]{@{}c@{}}Hate\\ Speech\end{tabular}}}       & Acc             & 78\%                   & 79.2\%          \\ \cline{2-4} 
                                                                                      & F1              & 0.77                   & 0.781           \\ \hline
\end{tabular}
\caption{Performance comparison between \textit{BERT\_Pre} and \textit{CoCNN+} in three downstream tasks}
\label{Tab: text_perform}
\end{table}
\subsubsection{Comparison in Downstream Tasks.}
We have compared \textit{BERT\_Pre} and \textit{CoCNN+} in three different downstream tasks: \\
(1) \textbf{Bengali Question Classification (QC):} This task consists of six classes (entity, numeric, human, location, description and abbreviation type question). The dataset has 3350 question samples \citep{Qdata}.\\
(2) \textbf{Hindi Product Review Classification:} The task is to classify a review into positive or negative class where the dataset consists of 2355 sample reviews \citep{data7}.\\
(3) \textbf{Hindi Hate Speech Detection:} The task is to identify whether a provided speech is a hate speech or not. The dataset consists of 3654 speeches \citep{data8}.

We use \textbf{five fold cross validation} while performing comparison on these datasets (see mean results in Table \ref{Tab: text_perform}) in terms of accuracy and F1 score). One tailed independent t-test with a P-value threshold of 0.05 have been performed on the 5 validation F1 scores obtained from five fold cross validation of each of the two models. Our \textbf{statistical test} results validate the significance of the improvement shown by \textit{CoCNN+} for all three of the mentioned tasks. 

\subsubsection{Vocabulary Learner as Spell Checker.}

\begin{table}[!htb]
\centering
\begin{tabular}{|c|c|c|}
\hline
\textbf{\begin{tabular}[c]{@{}c@{}}Spell Checker \\ Algorithm\end{tabular}} & \textbf{\begin{tabular}[c]{@{}c@{}}Synthetic\\ Error\end{tabular}} & \textbf{\begin{tabular}[c]{@{}c@{}}Real\\ Error\end{tabular}} \\ \hline
\textit{Vocabulary Learner}                                                & 71.1\%                                                             & 61.1\%                                                        \\ \hline
\textit{Phonetic Rule}                                                     & 61.5\%                                                             & 32.5\%                                                        \\ \hline
\textit{Clustering Rule}                                                   & 51.8\%                                                             & 43.8\%                                                        \\ \hline
\end{tabular}
\caption{Bengali spelling correction (accuracy)}
\label{fig: spell}
\end{table}

We also investigate the potential of \textit{VL} of \textit{CoCNN+} as a Bengali spell checker (SC). Both \textit{CoCNN} and \textit{CoCNN+} model use \textit{VL} for producing CNNvec representation from each input word. We extract the CNN sub-model of \textit{VL} from our trained (on Prothom-Alo dataset) \textit{CoCNN+} model. We produce CNNvec for all 13 K top words of Prothom-Alo dataset. For any error word, $W_e$, we can generate its CNNvec $V_e$ using \textit{VL}. We can calculate cosine similarity, $Cos_i$ between $V_e$ and CNNvec $V_i$ of each top word $W_i$. Higher cosine similarity means greater probability of being the correct version of $W_e$. We have discovered such approach to be effective for correct word generation. 
Recently, a phonetic rule based approach has been proposed by \citet{spell2}, where a hybrid of Soundex \citep{spell3} and Metaphone \citep{spell4} algorithm has been used for Bengali word level SC. Another SC proposed in recent time has taken a clustering based approach \citep{spell1}. We compare our proposed \textit{VL} based SC with these two existing SCs (see Table \ref{fig: spell}). Both the real and synthetic error dataset consist of 20k error words formed from the top 13k words of Prothom-Alo dataset. The real error dataset has been collected from a wide range of Bengali native speakers using an easy to use web app. Results show the superiority of our proposed SC over existing approaches.  

\section{Related Works}
Although a significant number of works for LM of high resource languages like English and Chinese are available, very few researches of significance for LM in low resource languages like Bengali and Hindi exist. In this section, we mainly summarize major LM related research works.

Sequence order information based statistical RNN models such as LSTM and GRU have been popular for LM tasks \citep{RNN1}. \citet{RNN3} showed the effectiveness of LSTM for English and French LM. The regularizing effect on LSTM was investigated by \citet{RNN4}. SOTA LSTM models learn sub-word information in each time stamp. \citet{LSTM2} proposed a morphological information oriented character N-gram based word vector representation. It was improved by \citet{LSTM1} and is known as FastText. \citet{LSTM3} proposed a technique for learning sub-word level information from data, while such idea was integrated in a character aware LSTM model by \citet{LSTM4}. \citet{LSTM_rel8} further improved word representation by combining ordinary word level and character-aware embedding. \citet{LSTM6} has shown that character-aware neural LMs outperform syllable-aware ones. \citet{LSTM5} evaluated such models on 50 morphologically rich languages.  

Self attention based Transformers have become the SOTA mechanism for sequence to sequence modeling in recent years \citep{tr2}. Some recent works have explored the use of such models in LM. Deep Transformer encoders outperform stacked LSTM models \citep{tr3}. A deep stacked Transformer model utilizing auxiliary loss was proposed by \citet{tr1} for character level language modeling. The multi-head self attention mechanism was replaced by a multi-linear attention mechanism with a view to improving LM performance and reducing parameter number \citep{tr_rel4}. Bengali and Hindi language, having unique characteristics, remain open as to what strategy to use for model development in such domains.

One dimensional version of CNNs have been used recently for text classification oriented tasks \citep{CNN_rel3,CNN_rel4,CNN_rel5}. \citet{CNN1} studied CNN application in LM showing the ability of CNNs to extract LM features at a high level of abstraction. Furthermore, dilated \textit{conv} was employed in Bengali LM with a view to solving long range dependency problem \citep{CNN2}.

\section{Conclusion}
We have proposed a CNN based architecture \textit{Coordinated CNN (CoCNN)} that introduces two 1D CNN based key concepts: word level \textit{VL} and sentence level \textit{TC}. Detailed investigation in three deep learning paradigms: CNN, LSTM and Transformer, shows the effectiveness of \textit{CoCNN} in Bengali and Hindi LM. We have also shown a simple but effective enhancement of \textit{CoCNN} by introducing skip connection and parallel \textit{conv} branches in the \textit{VL} and \textit{TC} portion, respectively. Future research may incorporate interesting ideas from existing SOTA 2D CNNs in \textit{CoCNN}. Over-parametrization and innovative scheme for \textit{CoCNN} pretraining are expected to increase its LM performance even further. %\emph{Codes and datasets of this research have been provided as supplementary materials. They will be made publicly available via GitHub upon request.} 

\section{Acknowledgements}
This work was supported by Bangladesh Information and Communication Technology (ICT) division [grant number: 56.00.0000.028.33.095.19- 85] as part of Enhancement of Bengali Language project.

\bibliography{main}
% \nobibliography{custom}

\end{document}